\documentclass[journal]{IEEEtran}
\usepackage{graphicx}
\usepackage{multirow}
\usepackage{amssymb}
\usepackage{amsmath,bm}
\usepackage{color}
\usepackage{amsthm}
\usepackage[ruled,vlined]{algorithm2e}
\usepackage{epstopdf}
\usepackage[colorlinks,linkcolor=black,anchorcolor=blue,citecolor=blue,urlcolor=blue]{hyperref}

\usepackage{ulem}
\usepackage{color}

\begin{document}
\title{Unified Multifaceted Feature Learning for Person Re-Identification}

\author{Cheng Yan\footnotemark{*}, Guansong Pang\footnotemark{*}, Xiao Bai, Chunhua Shen
\thanks{Cheng Yan and Xiao Bai are with School of Computer Science and Engineering, Beihang University, Beijing, China.}
\thanks{Guansong Pang and Chunhua Shen are with School of Computer Science, University of Adelaide.}
\thanks{* Equal Contribution.}

}

\maketitle

\begin{abstract}
Person re-identification (ReID) aims at re-identifying persons from different viewpoints across multiple cameras, of which it is of great importance to learn multifaceted features expressed in different parts of a person, e.g., clothes, bags, and other accessories in the main body, appearance in the head, and shoes in the foot. To learn such features, existing methods are focused on the striping-based approach that builds multi-branch neural networks to learn local features in each part of the identities, with one-branch network dedicated to one part. This results in complex models with a large number of parameters. To address this issue, this paper proposes to learn the multifaceted features in a simple unified single-branch neural network. The Unified Multifaceted Feature Learning (UMFL) framework is introduced to fulfill this goal, which consists of two key collaborative modules: compound batch image erasing (including batch constant erasing and random erasing) and hierarchical structured loss. The loss structures the augmented images resulted by the two types of image erasing in a two-level hierarchy and enforces multifaceted attention to different parts. As we show in the extensive experimental results on four benchmark person ReID datasets, despite the use of significantly simplified network structure, our method performs substantially better than state-of-the-art competing methods. Our method can also effectively generalize to vehicle ReID, achieving similar improvement on two vehicle ReID datasets.

\end{abstract}

\IEEEpeerreviewmaketitle

\section{Introduction}

Person re-identification (ReID) aims to search for the same person from a gallery of pedestrian images taken from different cameras, which is one critical task in computer vision due to its broad applications in domains such as multi-camera tracking, video surveillance, and forensic search. One significant challenge presented in person ReID is that, due to the possible shared face appearance, dressings or accessories among different persons, we need to examine the multifaceted features expressed in different parts of the detected persons to effectively perform the ReID task.

Many ReID methods have been introduced over the years, of which deep learning-based methods \cite{xiao2016learning,cheng2016person,sun2018pcb,cheng2016person,dai2019batch,huang2018adversarially,wang2018mancs,zheng2019joint} have gained significant improvement over traditional methods due to their remarkable ability to learn semantic-rich features. Particularly, the striping-based deep methods \cite{sun2018pcb,cheng2016person,wang2018learning,zhai2019defense,yao2019deep,hou2019interaction}, which builds multi-branch neural networks to learn local features in each of the predefined parts of the identities with one-branch network dedicated to one part, are introduced to learn the aforementioned multifaceted features. They are often the best performers on different benchmark datasets. However, these methods result in complex models, which involve a large number of parameters and computationally expensive training. They are therefore difficult to be well trained, especially when the given labeled training datasets are small.  

\begin{figure}[t]
\begin{minipage}[b]{1\linewidth}
  \centering
  \centerline{\includegraphics[width=1.00\linewidth]{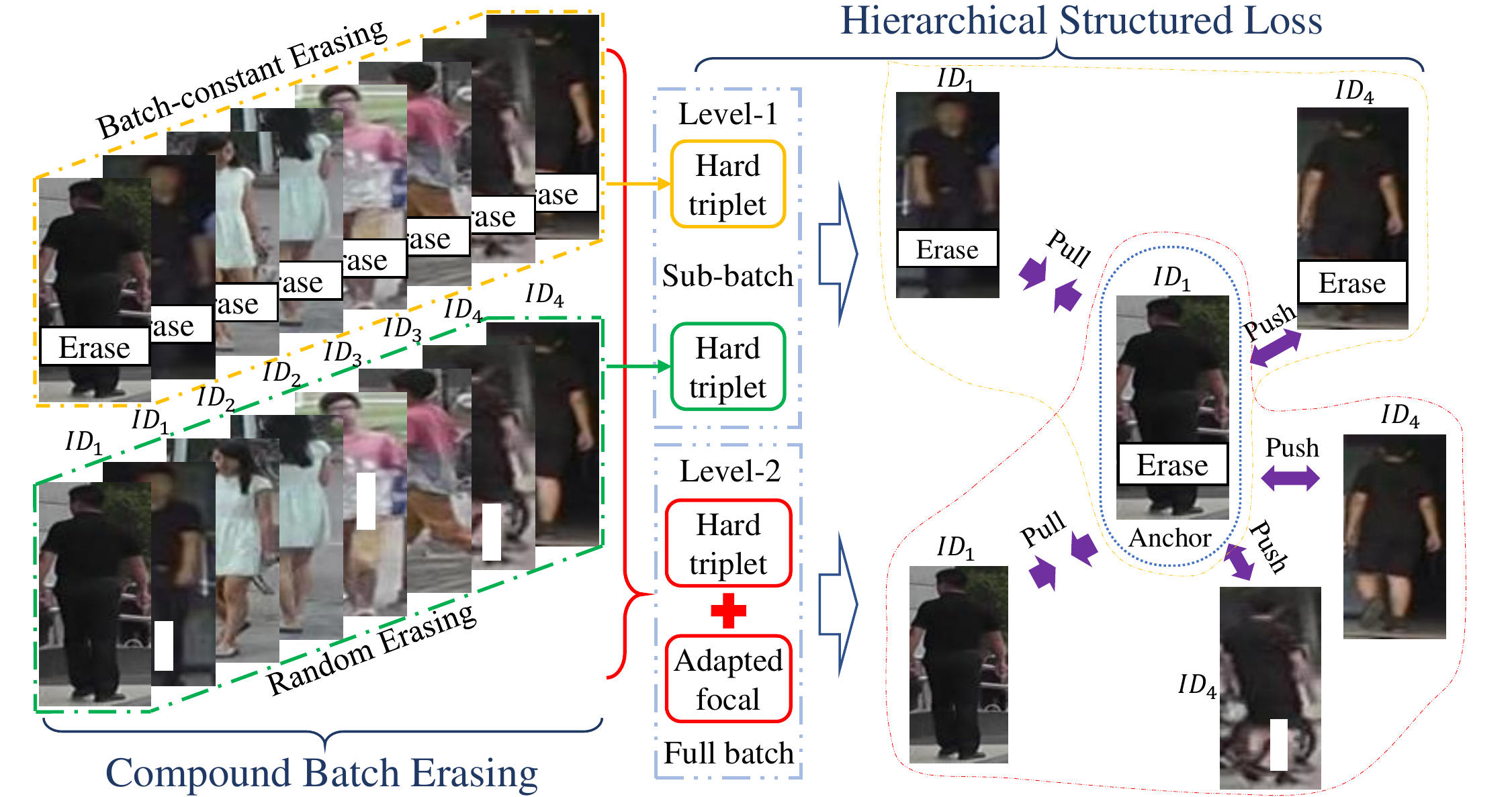}}
%  \vspace{1.5cm}
\end{minipage}
\caption{Two core modules of UMFL. Batch-constant erasing and random erasing are respectively applied two sub-batches with same images. A hierarchical structured loss is enforced on the sub-batches and the full batch to learn multifaceted features in an unified framework. Specifically, a hard-triplet loss is separately applied to the two sub-batches to learn within sub-batch features, while another hard-triplet loss and adapted focal loss are applied to the full batches to learn cross cross-sub-batches features}
\label{fig:example}
\end{figure}

To address this issue, in this work we propose to learn the multifaceted features in a simple unified singleton neural network. There have been a number of studies \cite{hermans2017defense,huang2018adversarially,wang2018mancs,luo2019bag,ristani2018features,shen2018deep} exploring the use of single network structure for the ReID task. However, their primary objective is to learn universally effective global features.  As a result, although they successfully learned discriminative global features, the models are typically attentive to only a single small discriminative part of the identity, leading to substantially less expressive features than that gained by striping methods. 
% Several attention mechanisms \cite{xu2018attention,chen2019abd,song2018mask,li2018harmonious} are introduced to learn more diverse attention, but they may have similar model complexity as the striping-based methods if complex attention mechanisms are used, and fail to learn multi-attention on different parts otherwise.

We introduce a novel framework, termed Unified Multifaceted Feature Learning (UMFL), to achieve the goal of learning complex multifaceted features with substantially simpler neural networks. The immediate gain is that, given the same amount of data, the resulting simplified neural networks can be trained significantly more effectively and efficiently than the complex ones. Specifically, as shown in Figure \ref{fig:example}, UMFL consists of two key collaborative modules, compound batch image erasing and hierarchical structured loss. The compound batch erasing is composed by two different types of image erasing operations, Batch-constant Erasing (BcE) and Random Erasing (RE), to generate the identity images with the chosen body part erased in all batch images (i.e., BcE) and random patches erased in each image of the batch (i.e., RE). RE \cite{zhong2017random} is now a key data augmentation ingredient and has shown a remarkable enabling power to improve the performance of person ReID \cite{luo2019bag,chen2019abd}, since it enables the model to focus on globally discriminative features. However, the RE-enabled models attend to single dominant discriminative image patches only. The use of the BcE-based augmented images enforces the models to search for non-erased body parts that are critical to person ReID. Since the erased body part is different in different batches, the ReID models are imposed to pay attention to the multifaceted features expressed in diverse discriminative body parts. Additionally, modeling with a mixture of BcE and RE-based augmented images help reinforce the learned multifaceted features.

To effectively learn the underlying multifaceted features in our compound batch erased images, we further introduce the hierarchical structured loss, which structures the BcE and RE-based augmented images into a two-level hierarchy per training batch and enforces separate losses to the augmented images in different groups of the hierarchy. Particularly, each of our training batch consists of two sub-batches, with one sub-batch containing the BcE augmented images and another sub-batch containing RE-based augmented images, resulting in three groups of image samples: BcE augmented sub-batch, RE augmented sub-batch, and the full batch. Separate losses are then applied to these three groups per batch to learn the multifaceted features.

Note that the trained UMFL-based single-branch network can use exactly the same network architecture as one branch of the multi-branch complex networks that aim to learn global features. Therefore, the UMFL model can also be applied as a base block to plug into the complex networks to improve their performance.

Additionally, UMFL is very different from the recently proposed batch dropblock network (BDB) \cite{dai2019batch}. The batch-constant block dropping in BDB is similar to BcE in UMFL, but they are actually two different operations since BDB performs the dropping in the feature map layer while UMFL applies BcE to the original image. This is the only similarity between BDB and UMFL. UMFL and BDB use completely different approaches (single-branch network with hierarchical structured loss vs. multi-branch networks with single loss). As a result, they have extremely different properties. For example, BDB involves substantially more parameters and it is much more difficult to train than UMFL; UMFL is generic and can plug into other types of approaches, whereas BDB does not have this flexibility. As we show in our experiments, UMFL can easily plug into BDB and other recent state-of-the-art methods such as ABD \cite{chen2019abd} to achieve new performance benchmarks. In summary, this work makes the following three main contributions.

\begin{itemize}
    \item We propose a novel framework termed Unified Multifaceted Feature Learning to learn diverse discriminative features expressed in different parts of the identifies using single-branch neural networks. As we show in our experiments on four benchmark datasets, despite the use of significantly simplified network architectures, the resulting model is able to capture multifaceted features that can often be captured by state-of-the-art complex multi-branch networks only.
    
    % that combines data augmentation with hierarchical loss frame to improve CNN's generalization ability for ReID task.
    \item Our UMFL model is generic and can be easily incorporated into state-of-the-art complex methods to substantially improve their performance.
    
    \item Beyond person ReID, UMFL can also generalize to other similar tasks and achieve state-of-the-art performance on two vehicle ReID benchmark datasets.
\end{itemize}

% Extensive experimental results on four benchmark person ReID and two vehicle ReID datasets show that the proposed UMFL frame improves the performance in several types of state-of-the-art approaches.

% \begin{table*}[htbp]
% \centering
% \caption{Comparison between Hard-mining triplet loss and ours. RE refers to Random Erasing and BcE refers to Batch-constant Erasing}

% \scalebox{0.85}{
% \begin{tabular}{c|c|c}
% \hline
% & \textbf{Random Erasing}  & \textbf{Ours}\\
% \hline
% Erasing type & Random Erasing(RE) with 0.5 probability &  Batch-constant Erasing(BcE) with random location \\
% \hline
% Batch size & B &  B*2 \\
% \cline{2-3}
% (Details) & B samples with RE &  B samples with RE and B samples with BcE\\
% \hline
% Triplet loss from & 1 group: &  3 group: \\
% \cline{2-3}
% (Details) & all samples & all samples, RE group, BcE group \\
% \hline
% Focus & Beyond random area of different images & Beyond same area of different images  \\
% \hline
% \end{tabular}
% }
% \label{tab:tl_hap}
% \end{table*}

\section{Related Work}
Due to the remarkable capacity of feature learning of CNN, many deep learning based ReID methods \cite{xiao2016learning,cheng2016person,sun2018pcb,cheng2016person,dai2019batch,huang2018adversarially,wang2018mancs,zheng2019joint} have been proposed recently \cite{das2014consistent,li2013learning,liao2015person,ma2013domain,pedagadi2013local,zheng2012reid}. These methods can be divided into supervised and unsupervised approaches. Since our method uses labels for training, we only review the supervised methods.

\textbf{Multifaceted Feature Learning}. In supervised ReID methods, the majority of these deep learning based methods focus on designing network structures to divide the image into several stripes and learn local features within each stripe \cite{sun2018pcb,cheng2016person,dai2019batch,wang2018learning,zhai2019defense,yao2019deep,hou2019interaction}. They enforce the learner to pay more attention to different parts of the identities by combining the striping local features. At the testing stage, different part features and global features are concatenated together as the final representation. This is one of the most effective approaches, but their networks are usually complex, and thus they are computationally expensive and often difficult to be trained well. Some attention-aware methods \cite{li2018harmonious,xu2018attention} are proposed to enhance the attentiveness of CNN which also benefit from the complex striping structure.

\textbf{Data Augmentation}. The other methods focus on data augmentation or loss function. Data augmentation includes GAN based approaches which using GAN to generate more data for training \cite{wei2018person,zhong2018camstyle}, mask or pose guided framework \cite{kalayeh2018human,saquib2018pose} that utilizes extra semantic information from pose estimation or segmentation models, as well as  random erasing approach that randomly erases a small area of input images \cite{zhong2017random}. Some other data augmentation studies \cite{zhang2017mixup,inoue2018data,verma2018manifold} have shown that, combinations of examples and labels of training data can also effectively regularize the neural network. Random erasing uses no extra information and it is arguably the simplest effective method. 
% , however, the linear combinations only works well for classification.
% The GAN, mask or pose based methods use other networks to generate image to increase the number of input images or improve the mask of input for augmentation, however, the benefit comes from the help of other network with extra semantic information. On the contrary, r
\begin{figure*}[htbp]
\begin{minipage}[b]{1\linewidth}
  \centering
  \centerline{\includegraphics[width=0.9\linewidth]{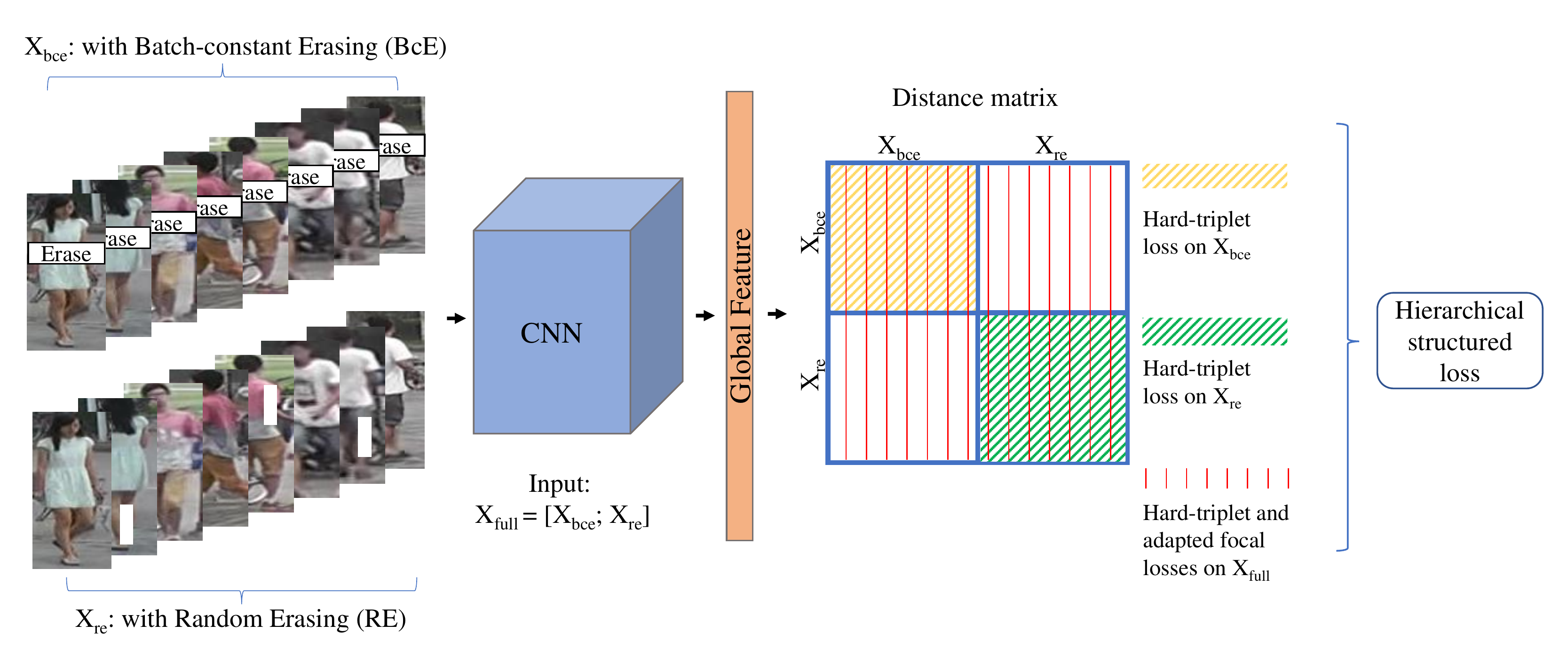}}
%  \vspace{1.5cm}
\end{minipage}
\caption{An overview of the Unified Multifaceted Feature Learning framework. Specifically, the compound batch erasing operation is first applied to the input images of two sub-batches $\mathbf{X}_{\textit{re}}$ and $\mathbf{X}_{\textit{bce}}$ that are made up with the same images but having RE and BcE erasing operation respectively. These two sub-batches are concatenated into the full batch $\mathbf{X}_{\textit{full}} = [\mathbf{X}_{\textit{re}}; \mathbf{X}_{\textit{bce}}]$. These hierarchically organized batch samples are fed into CNN for feature learning in which any deep CNN network backbone can be used. At the end of the network, the proposed hierarchical structured loss, including hard-triplet loss separately applied to $\mathbf{X}_{\textit{re}}$, $\mathbf{X}_{\textit{bce}}$ and $\mathbf{X}_{\textit{full}}$ as well as an adapted focal loss on $\mathbf{X}_{\textit{full}}$, together with a commonly used classification loss, are applied to drive the CNN network to learn diverse and discriminative features from different parts of the identities.}
\label{fig:frame}
\end{figure*}

\textbf{Loss Function}. The triplet loss is now probably the most popular loss used in current state-of-the-art ReID methods. Some advanced versions have been proposed such as batch-hard triplet mining \cite{hermans2017defense} and batch-soft triplet mining \cite{ristani2018features}. In batch-hard triplet loss, only the most hard positive and negative samples are selected for each anchor to form triplet. In batch-soft triplet loss, the triplets are re-weighted to prevent the influence from outlier samples. Some other loss functions also have been used for boosting the performance such as quadruplet loss \cite{chen2017beyond} in which quadruplet networks with quadruplet sampling are used for training.

\section{Unified Multifaceted Feature Learning}

\subsection{The Proposed Framework}
In a person ReID system, given a set of training images $\mathcal{X}=\{\mathbf{x}_{1},\cdots, \mathbf{x}_{N}\}$ and the corresponding one-hot encoding of the identity/class set $\mathcal{Y}=\{\mathbf{y}_{1},..., \mathbf{y}_{N}\}$, then the task is to learn a mapping function $\phi: \mathcal{X} \times \mathcal{Y} \mapsto \mathcal{Z}$ which projects the original data $\mathcal{X}$ onto a new feature space $\mathcal{Z}$, such that the distance of the images of each person is small while the distance w.r.t different persons is large. Given a query image $\mathbf{q}$, the system first computes the distance between $\phi(\mathbf{q})$ and each image from a gallery image set $\mathcal{G}=\{\mathbf{g}_{1},\cdots, \mathbf{g}_{M}\}$, and then returns the images that have the smallest distance to the query image. Note that $\mathcal{X}$ and $\mathcal{G}$ are normally assumed to have no overlapping identities.
% The ReID task is considered as a zero-shot learning problem since the identities in gallery set and the training set have no overlapping.

Our proposed Unified Multifaceted Feature Learning (UMFL) framework leverages two collaborative modules, compound batch erasing and hierarchical structured loss, to learn the diverse features expressed in different parts of persons using a simple single-branch neural network. They are collaborative in the sense that the synthesis of these two modules in UMFL works substantially better than that with one of these two modules replaced with other functions. The procedure of UMFL is presented in Figure \ref{fig:frame}. Specifically, the compound batch erasing first generates a structured batch with two sub-batches, one sub-batch samples generated by BcE and another sub-batch samples generated by RE. The samples are then projected onto the $\mathcal{Z}$ space by a single-branch network architecture, in which the hierarchical structured loss is applied to the each batch samples that are structured into a two-level hierarchy (i.e., two sub-batches in the first level and the full batch in the second level) to learn the multifaceted features carried by the sub-batch and cross sub-batch samples.

% As shown in Figure \ref{fig:frame}, it first augments each input batch data by using general augmentation and batch-constant erasing strategies. Then it employs a convolutional neural network (CNN) to extract features, in which any CNN architecture can by used as the backbone. Finally, the network is restrained by the proposed hierarchical loss function. The whole frame aims to improve the generalization ability of CNN for ReID tasks.

% We first give the problem formulation of ReID task. Our data augmentation strategy and our loss function will be presented in Section \ref{sec:augmentation} and Section \ref{sec:loss} respectively.

\subsection{Compound Batch Erasing}
\label{sec:augmentation}
Compound batch erasing is composed by the widely-used random erasing (RE) and our proposed batch-constant erasing (BcE), which generates two sub-batches of augmented identity images. Specifically, for each batch, its two sub-batches contain exactly the same raw image set but are respectively applied with RE and BcE to generate sub-batch images with different erased areas. This is to have more effective learning of the multifaceted features using a fixed number of identities. To achieve that, we first sample a sub-batch of images $\mathbf{X} \in \mathbb{R}^{B \times 3 \times H \times W}$, we then duplicate the sub-batch and concatenate the two sub-batches to form the full batch $\mathbf{X}_{\textit{full}} = [\mathbf{X}_{\textit{re}}; \mathbf{X}_{\textit{bce}}] \in \mathbb{R}^{2B \times 3 \times H \times W}$, of which RE and BcE are then respectively applied to $\mathbf{X}_{\textit{re}}$ and $\mathbf{X}_{\textit{bce}}$. 
% RE and BcE are then respectively applied to each sub-batch to generate sub-batch images with different erased areas.
% Motivated by mixup strategy \cite{zhang2017mixup,inoue2018data,verma2018manifold} and considering the particularity of ReID task in which the testing is based on the feature of each image. We propose a batch augmentation method named Batch Compound Augmentation, in which we double enlarge one batch by using same images with different augmentation strategies. This batch augmentation acts as a regularizer, increasing both generalization and performance scaling.  

\subsubsection{Random Erasing (RE)}
% Batch Compound Augmentation (BCA) is applied on a batch in which two sub-batch of same images are augmented with two different strategies: Batch-constant Erasing (BcE) and General Augmentation (GA). For each batch, BcE is applied on all the images of one sub-batch with erasing on the same area at a random spacial location. GA is applied on the other sub-batch in which images are randomly augmented with common used augmentation strategy, such as random cropping, random flipping and random erasing. Specifically, 

By randomly erasing a small rectangle patch of training identity images, the RE-based augmentation substantially improves ReID methods in learn discriminative features \cite{luo2019bag}. RE is extremely simple and works as follows.  Particularly, for each image in a given sub-batch $\mathbf{X}_{\textit{re}}$, a 50\% chance is set to randomly erase a $a \times b$ rectangle area for erasing, where $a = \sqrt{\frac{w \times h }{r}}$ and $b = \sqrt{w \times h  \times r}$ ($\frac{w \times h}{W \times H}$ is predefined in the range of $[s_{l}, s_{h}]$ and is used as the area ratio; $r \in [r_{1}, r_{2}]$ is an aspect ratio; the hyperparameters $s_{l}$, $s_{h}$, $r_{1}$ and $r_{2}$ are used with the recommended settings in \cite{luo2019bag}).

% , where we set $\frac{w \times h}{W \times H}$ in the range of $[s_{l}, s_{h}]$ considered as the area ratio, and an aspect ratio $r \in [r_{1}, r_{2}]$ is set as the aspect ratio. The $w$ and $h$ of the rectangle area is set to $\sqrt{\frac{w \times h }{r}}$ and $\sqrt{w \times h  \times r}$ respectively. Finally, the areas $\textit{area}$ in all images of the $\mathbf{\bar{X}}$ are erased with mean value of dataset.

\subsubsection{Batch-constant Erasing (BcE)}
Different from the random erasing or other deterministic erasing methods \cite{singh2017hide,devries2017improved} that erase a small rectangle area only and the erasing area is randomly chosen per image, our batch-constant erasing method erases a striping part of the image and the erased part is fixed and applied to all the images in the same sub-batch $\mathbf{X}_{\textit{bce}}$. The erased part per sub-batch is randomly chosen. More specifically, we first spatially divide the image into $s$ horizontal parts, and then randomly choose one part and erase the same part of all the images per sub-batch $\mathbf{X}_{\textit{bce}}$. Applying a proper loss to the resulting sub-batches with different constant erased parts would effectively force the model to learn the similarities and differences of the non-erased parts, resulting in diverse attention to different parts of the identities.

\subsection{Hierarchical  Structured Loss}
\label{sec:loss}
Separate losses are then applied to the two sub-batches and the full batch to learn three types of features, including the globally discriminative features from the sub-batch $\mathbf{X}_{\textit{re}}$, diverse discriminative features in different parts from the sub-batch $\mathbf{X}_{\textit{bce}}$, global and part mixed features from the full batch $\mathbf{X_{\textit{full}}}$. These separate losses are termed hierarchical structured loss to emphasize the importance of using the two-level hierarchical structure of the batch and the corresponding applied losses, since we cannot capture such rich features otherwise. Below we introduce the detail of each separate loss.
\subsubsection{Hard-triplet Loss on sub-batches}\label{subsec:sub-batch}

Triplet loss shows remarkable enabling power in current state-of-the-art ReID methods and is now probably the most widely used loss in the ReID task. It takes three samples, including an anchor $\mathbf{x}_{a}$, a positive sample $\mathbf{x}_{p}$ that comes from the same person as $\mathbf{x}_{a}$, and a negative sample $\mathbf{x}_{n}$ taken from a person different from that of the anchor, as a triplet to learn feature representations. In the learning process, it enforces the inter-person distances are greater than the intra-person distances by at least a predefined margin $m$. The generic triplet loss is defined as follows:
\begin{equation}
\label{triplet}
L_{t}=  [d(\mathbf{z}_{a}, \mathbf{z}_{p})-d(\mathbf{z}_{a}, \mathbf{z}_{n}) + m]_{+},
\end{equation}
where $\mathbf{z}=\phi(\mathbf{x})$ denotes the learned feature representation of $\mathbf{x}$, $d(\cdot,\cdot)$ is the distance of two samples,  $m$ is a predefined margin and $[\cdot]_{+}$ represents $\max(\cdot, 0)$. Convolutional networks are often employed to instantiate the $\phi$ function. However, the vanilla triplet loss in Eqn.~(\ref{triplet}) works well only when carefully selected triplets are provided, which is often impossible to perform at scale. An advanced triplet loss known as the hard triplet mining \cite{liao2015efficient,su2016deep,liu2017end} are used to deal with the problem, which are defined as:

\begin{equation}
\label{triplet_hard}
L_{ht} =  \left[\max_{p \in \mathcal{P}_{a}} d(\mathbf{z}_{a}, \mathbf{z}_{p})- \min_{n \in \mathcal{N}_{a}} d(\mathbf{z}_{a}, \mathbf{z}_{n}) + m\right]_{+},
\end{equation}
where $\mathcal{P}_{a}$ ($\mathcal{N}_{a}$) is the collection of positive (negative) samples of the anchor in a batch. This loss enables better results than the original triplet loss \cite{schroff2015facenet}. The role of the hinge function $[\cdot]_{+}$ is to avoid correcting `already correct' triplets, but, as shown in \cite{hermans2017defense}, replacing the hinge function with the softplus function $\log\left(1 + \exp(\cdot)\right)$ can work in a similar way while at the same time avoiding the use of the hard cut-off margin $m$. So we adopt the softplus function and obtain the following hard-triplet loss:
\begin{equation}
\label{triplet_hard_soft}
L_{\textit{sht}} =  \log \Big(1 + \exp\big(\max_{p \in \mathcal{P}_{a}} d(\mathbf{z}_{a}, \mathbf{z}_{p})- \min_{n \in \mathcal{N}_{a}} d(\mathbf{z}_{a}, \mathbf{z}_{n})\big)\Big).
\end{equation}

The obtained hard-triplet loss is applied to the two sub-batches $\mathbf{X}_{\textit{re}}$ and $\mathbf{X}_{\textit{bce}}$ separately:

\begin{equation}
\label{h-hard-triplet}
L_{\textit{sht}}^{\textit{sub}} = L_{\textit{sht}}(\mathbf{X}_{\textit{re}})+L_{\textit{sht}}(\mathbf{X}_{\textit{bce}}),
\end{equation}
where $L_{\textit{sht}}(\mathbf{X}_{*})$ means $L_{\textit{sht}}$ is applied to the sub-batch $\mathbf{X}_{*}$.

% After compound batch erasing, half of images from each batch are erased at the same area leading to the increase of hard-triplet samples. 
The term $L_{\textit{sht}}(\mathbf{X}_{\textit{bce}})$ enables the learning of discriminative multifaceted features from diverse parts. Intuitively, for a sub-batch of images samples with a highly discriminative part removed in $\mathbf{X}_{\textit{bce}}$, minimizing $L_{\textit{sht}}(\mathbf{X}_{\textit{bce}})$ forces the model to learn features from other discriminative parts. For example, as shown in Figure \ref{fig:example}, the anchor image sample is from an identity $\textit{ID}_{1}$, and the black trouser of this identify is one of the most discriminative parts to differentiate images of $\textit{ID}_{1}$ from that of $\textit{ID}_{4}$ who exhibits extremely similar appearance to $\textit{ID}_{1}$ except the black trouser part. When presenting $\mathbf{X}_{\textit{bce}}$ with the black trouser part erased, enforcing $L_{\textit{sht}}(\mathbf{X}_{\textit{bce}})$ on this sub-batch teaches the model attend to other discriminative information in other parts such as the head or body part.

However, using $L_{\textit{sht}}(\mathbf{X}_{\textit{bce}})$ only may result in the loss of globally discriminative features since a large fixed striping part is always blocked in each sub-batch $\mathbf{X}_{\textit{bce}}$. Therefore, we apply $L_{\textit{sht}}(\mathbf{X}_{\textit{re}})$ to the other sub-batch $\mathbf{X}_{\textit{re}}$. Since only a small random chosen area is erased and the erased area is often different per image in $\mathbf{X}_{\textit{re}}$, imposing $L_{\textit{sht}}(\mathbf{X}_{\textit{re}})$ to $\mathbf{X}_{\textit{re}}$ can complement $L_{\textit{sht}}(\mathbf{X}_{\textit{bce}})$ in learning globally discriminative features.

% and since the most discriminative part is erased, the images from $ID_{4}$ has a very similar appearance with anchor. As a result, the feature of this image from $ID_{4}$ is more close to anchor than that of some images from $ID_{1}$. Single hard triplet for the whole batch is insufficient to guarantee the feature learning. Here we use two level hierarchical hard-triplet loss, defined as:
% \begin{equation}
% \label{h-hard-triplet}
% L_{hht} = \sum_{i=1}^{3} L_{\textit{sht}}^{batch_{i}}.
% \end{equation}

% This loss is made up by three parts of batch hard-triplet loss acting on sub-batch $X_{\textit{bce}}$, $X_{\textit{re}}$ and the full batch $X_{full}$. The first one keep  images with same spacial erasing area meet the triplet constraint, which is the driving force of generalization because it makes the network pay more attention to different parts of images. The second part guarantees the individually-differentiated images to satisfy triplet constraint. The last part gives an assurance on the full batch image meeting the triplet constraint.

\subsubsection{Triplet and Focal Losses on the Full Batch}

$L_{\textit{sht}}(\mathbf{X}_{\textit{bce}})$ and $L_{\textit{sht}}(\mathbf{X}_{\textit{re}})$ are focused on learning features carried by the erased images within each sub-batch, but they cannot capture the features in the same/different identities across the full batch $\mathbf{X}_{\textit{full}}$. We therefore apply two additional losses, including the hard-triplet loss and an adapted focal loss, to the full batch to learn such features. They are added to learn finer-grained discriminative features (via the hard-triplet loss) and additional important features (via the adapted focal loss).

Specifically, similar to $\mathbf{X}_{\textit{bce}}$ and $\mathbf{X}_{\textit{re}}$, the hard-triplet loss is also applied to $\mathbf{X}_{\textit{full}}$ as:
\begin{equation}\label{sht_full}
    L_{\textit{sht}}^{\textit{full}} = L_{\textit{sht}}(\mathbf{X}_{\textit{full}}).
\end{equation}
This helps learn finer-grained discriminative features than that from the two sub-batches as it is applied to images with more areas erased.

Focal loss is enforced to learn from hard negative samples, i.e., image pairs that are from different identities and become exceedingly similar due to the compound batch erasing. This is due to the fact that in the full batch $\mathbf{X}_{\textit{full}}$, each image has two versions, one with a striping part erased and another with a small random area erased, and as a result, the erased negative images, especially the one with the striping part erased, may become undistinguished from the erased anchor images, leading to the decrease of distance between the negative samples and the anchor samples. It is difficult for the hard-triplet loss to cope with the problem, because our erasing operation can also result in large distances between images of the same identity. Therefore, using the distance between the anchor and positive samples to guide the penalization on the negative samples may be misleading. An adapted focal loss is therefore introduced to address this issue. 

Focal loss is a dynamically scaled cross entropy loss, in which a scaling factor can automatically down-weight the contribution of easy examples during training , enabling the model to rapidly focus on hard examples \cite{lin2017focal}. The original focal loss is defined as follows.
\begin{equation}
\label{focal}
L_{f}^{\textit{full}} = \sum-(1-p)^{\gamma} \log p,
\end{equation}
where $p$ is the probability and $\gamma$ is the scaling factor. It automatically reduces the importance of the samples having small $p$. To adapt the focal loss to our task, an adaptive sigmoid based loss function is defined as follows to transform pairwise distances between identity image samples to probabilities:
\begin{equation}
\label{p}
p = \frac{2}{(1+e^{-\alpha d(\mathbf{z}_{i}, \mathbf{z}_{j})})} - 1,
\end{equation}
where $d(\mathbf{z}_{i}, \mathbf{z}_{j})$ is the distance between samples $\mathbf{x}_{i}$ and $\mathbf{x}_{j}$ in the representation space. Eq. (\ref{p}) is defined to transform the range $p \in [0, +\infty)$ in the original adaptive sigmoid function $1/(1+e^{-\alpha d})$ \cite{wang2013nonlinear} to $p \in [0, 1)$. 
% $d \in [0, +\infty)$

% Applying the focal loss $L_{f}$ with the adapted probability function in Eq. (\ref{p}) guarantees that large penalization is always enforced on the erased negative samples and anchor samples if they have small distances, and the penalization vanishes when they have sufficiently large distances. This guarantees enables our model to effectively learn from hard negative samples that the triplet loss fails.

% With the hard negative samples increase, only using full batch hard-triplet as the whole batch constraint is not enough to guarantee the relationship of samples in which the inter-person distance should be large enough than intra-person distance. To overcome this problem, we propose a focal-loss based pairwise loss function to restrain negative samples which will be introduced in next sub-section.

% In our frame, we want the distance between negative sample and anchor large and stop punishment when it is large enough. However, triplet loss can not give punishment once the distance of negative pair is large than that of the positive pair. So we adopt focal-loss based pairwise loss function to give each negative pair a corresponding punishment. 

Note that the incorporation of the focal loss does not introduce any extra computation of the distances, because the distances used in Eq. (\ref{sht_full}) can be directly used in Eq. (\ref{p}).

Overall, our model is driven by the collaboration of the following four losses.

\begin{equation}
\label{loss}
L = L_{\textit{sht}}^{\textit{sub}} + L_{\textit{sht}}^{\textit{full}} + L_{f}^{\textit{full}} + L_{c},
\end{equation}
where $L_{c} = \mathbb{E} (\bar{\mathbf{y}}_{i}, \mathbf{y}_{i})$ is a classification loss based on cross entropy between the prediction $\bar{\mathbf{y}}_{i}$ and the ground truth $\mathbf{y}_{i}$ as in most ReID methods \cite{dai2019batch,huang2018adversarially,wang2018mancs,zheng2019joint}.
% Finally, similar to most ReID methods \cite{dai2019batch,huang2018adversarially,wang2018mancs,zheng2019joint}, a classification loss is added to the loss function:
% \begin{equation}
% \label{cla}
% L_{c} = \mathbb{E} (\mathbf{z}_{i}^\intercal \mathbf{W}, \mathbf{y}_{i}),
% \end{equation}
% where $\mathbf{W}$ is the classifier. The final loss for each anchor is:

\section{Experiments}

\renewcommand{\arraystretch}{0.9}
\begin{table*}[htbp]
\centering
\caption{MAP and R-1 of different methods on four benchmark datasets. The upper part shows the results of global feature-based methods. The lower part shows the results of striping methods. The best performance per group is boldfaced.}

\begin{tabular}{c|c|c c|c c|c c|c c}
\hline
\hline
\multirow{3}{*}{\textbf{Methods}} &\multirow{3}{*}{\textbf{Source}} & \multicolumn{4}{ |c }{\textbf{CUHK03}} &  \multicolumn{2}{ |c| }{\multirow{2}{*}{\textbf{Market1501}}} & \multicolumn{2}{ |c }{\multirow{2}{*}{\textbf{DukeMTMC}}}\\
\cline{3-6}
& &  \multicolumn{2}{ |c| }{Detected} & \multicolumn{2}{ |c| }{Labeled} &  &  &  &  \\
\cline{3-10}
& & mAP & R-1 & mAP & R-1 & mAP & R-1 & mAP & R-1 \\
\hline
% \multirow{4}*{\shortstack{Data\\Augumentation}}
% CAN \cite{liu2017end}& TIP2017  & -& -& -& - & 35.9& 60.3 & - & -  \\
% TriNet \cite{hermans2017defense}& Arxiv17  & -& -& -& - & 69.1& 84.9 & - & -  \\
AWTL \cite{ristani2018features}& CVPR18  & - & - & - & - & 75.7 & 89.5 & 63.4 & 79.8 \\
AOS \cite{huang2018adversarially}& CVPR18  & 43.3  & 47.1 & -  & - & 70.4  & 86.4  & 62.1  & 79.1  \\
GSRW \cite{shen2018deep}& CVPR18 &  - & - & -  & - & 82.5  & 92.7  & 66.4  & 80.7 \\ 
Mancs \cite{wang2018mancs}& ECCV18  & 60.5 & 65.5 & 63.9 & 69.0  & 82.3  & 93.1  & 71.8  & 84.9 \\
% SVDNet \cite{sun2017svdnet}  & ICCV17 & 37.3 & 41.5 & 37.8 & 40.9&  62.1 & 82.3 & 56.8 & 76.7  \\
Camstyle \cite{zhong2018camstyle}& TIP18  &  - & - & - & -& 68.7 & 88.1 & 53.5 & 75.3 \\
SPReID \cite{kalayeh2018human} & CVPR18  & -&-&-&-& 81.3 & 92.5 & 71.0 & 84.4 \\
PN-GAN \cite{qian2018pose} & ECCV18 & - & - & - & -& 72.6 & 89.4 & 53.2 & 73.6 \\
VCFL \cite{liu2019view}& ICCV19  &  55.6 & 61.4 & -  & - & 74.5  & 89.3  & -  & - \\
\texttt{ASB} \cite{luo2019bag} & CVPRW19 & 58.2 & 60.5 & 60.2 & 62.1  & 85.9  & 94.5  & 76.4  & 86.4 \\
\texttt{UMFL-enabled ASB} & Ours & \textbf{68.7} & \textbf{71.1} & \textbf{71.9} & \textbf{73.9} & \textbf{87.5} & \textbf{94.8} & \textbf{77.2} &\textbf{88.0}  \\
\hline
\hline
% & Baseline1 \cite{luo2019bag} & 52.5 & 54.2 & 55.3 & 56.3 & 82.3 & 93.5 & 71.0 & 84.9   \\
% & Ours + Baseline1  & \textbf{63.5} & \textbf{66.1} & \textbf{67.1} & \textbf{69.1} &  \textbf{84.5} & \textbf{94.2} & \textbf{73.4} & \textbf{86.0}  \\
% AlignedReID \cite{zhang2017alignedreid}& Arxiv2017  &  - & - & - & - & 77.7 & 90.6 & 67.4 & 81.2 \\
MLFN \cite{chang2018multi}& CVPR18  & 47.8  & 52.8 & 49.2  & 54.7 & 70.4  & 86.4  & 62.1  & 79.1  \\
HA-CNN \cite{li2018harmonious}& CVPR18  & 38.6  & 41.7 & 41.0  & 44.4 & 75.7  & 91.2  & 63.8  & 80.5  \\
AACN \cite{xu2018attention}& CVPR18  & -  & - & -  & - & 83.0  & 88.7  & -  & - \\ 
MGCAM \cite{song2018mask}& CVPR18 &  46.7 & 46.8 & 50.1  & 50.2 & 74.3  & 83.8  & 66.4  & 80.7 \\ 
PCB \cite{sun2018pcb}& ECCV18   &  - & - & -  & - & 77.4  & 92.3  & 65.3  & 81.9 \\
PL-Net \cite{yao2019deep} & TIP19 &  - & - & -  & - & 69.3  & 88.2  & -  & - \\
IANet \cite{hou2019interaction}& CVPR19  &  - & - & -  & - & 83.1  & 94.4  & 73.4  & 87.1 \\
MCG \cite{zhai2019defense}& CVPRW19  &  - & - & 55.3  & 61.7 & 78.3  & 92.6  & 69.4  & 84.7 \\
\texttt{BDB} \cite{dai2019batch}& ICCV19  & 69.3 & 72.8 & 71.7  & 73.6 & 82.8  & 93.5  & 71.5  & 86.8 \\
\texttt{UMFL-enabled BDB} & Ours & 69.0 & 72.6 & 72.0 & 75.0 & 84.5 & 94.6 & 73.5 & 87.3 \\
\texttt{ABD} \cite{chen2019abd}& ICCV19   & 66.8 & 69.5  & 69.5  & 72.7 & 88.2  & 95.6 & 78.5  & 89.0\\
\texttt{UMFL-enabled ABD} & Ours &\textbf{70.1} & \textbf{73.7} & \textbf{73.2} & \textbf{75.6}  & \textbf{88.7}  & \textbf{95.7} & \textbf{79.3} & \textbf{89.5}\\
% \hline
% Ours + ASB &  & 68.7 & 71.1 & 71.9 & 73.9 & 87.5 & 94.8 & 77.2 & 88.0  \\
\hline
\hline
\end{tabular}
\label{tab:MAP}
\end{table*}

\subsection{Datasets}
We evaluate the performance on four widely used person ReID datasets from CUHK03 \cite{li2014deepreid}, Market1501 \cite{zheng2015scalable}, DukeMTMC-ReID \cite{zheng2017unlabeled}. 

CUHK03 contains the image set with 14,096 images from 1,467 identities captured from six cameras in the CUHK campus and each person only has two camera views. Following \cite{zhong2017re}, we applied the CUHK03-NP splits, in which 767 identities and the other 700 ones are selected for training and testing respectively. For this dataset, two types of datasets are built based on the way of creating bounding boxes. Specifically, CUHK03-Detected uses pedestrian detectors to create the bounding boxes while that of CUHK03-Labeled is manually labeled. 
%The pedestrian detector-based method is more challenging than the manually labeled one since the former is less accurate. 

Market1501 is a large person ReID dataset containing 12,936 images from 751 identities in the training data, and 3,368 query images and 19,732 gallery images from 750 identities in the testing data. These images were captured from 6 different camera viewpoints with manual bounding boxes. 
%There are about 17 images for each identity.
% The identities in the training and testing sets have no overlapping.

DukeMTMC-ReID is a subset of DukeMTMC \cite{ristani2016performance} for person ReID. The images are cropped by hand-drawn bounding boxes. The data was taken from 8 cameras of 1,404 identities with respective 16,522, 2,228 and 17,661 images in the training, query and gallery sets. 
% There is no overlapping in the training and testing sets.

\subsection{Evaluation Protocol} 
Following the standard protocol in \cite{song2018mask,wang2018mancs,sun2018pcb,chen2018person,sun2019perceive}, we use Cumulated Matching Characteristics (CMC) and  mean average precision (mAP) to evaluate the performance on all datasets. We report the cumulated matching accuracy at rank 1 (R-1 for short) and the mAP value of the retrieval performance. Note that all the reported results here do not involve re-ranking, though it may be used as an extra step to further improve the accuracy.

\renewcommand{\arraystretch}{0.9}
\begin{table*}[htbp]
\centering
\caption{MAP and R-1 of the UMFL-enabled ASB and its ablated variants.}

%\scalebox{1.0}{
\begin{tabular}{l|c c|c c|c c|c c}
\hline
\multirow{3}{*}{\textbf{Methods}} & \multicolumn{4}{ |c }{\textbf{CUHK03}} &  \multicolumn{2}{ |c| }{\multirow{2}{*}{\textbf{Market1501}}} & \multicolumn{2}{ |c }{\multirow{2}{*}{\textbf{DukeMTMC}}}\\
\cline{2-5}
&  \multicolumn{2}{ |c| }{Detected} & \multicolumn{2}{ |c| }{Labeled} &  &  &  &  \\
\cline{2-9}
& mAP & R-1 & mAP & R-1 & mAP & R-1 & mAP & R-1 \\
\hline
Base + hard-triplet & 52.1 & 54.2 & 55.4 & 56.1 & 82.3 & 93.0 & 71.0 & 84.9  \\
Base + RE + hard-triplet (ASB) & 58.3 & 60.1 & 59.1 & 62.1 & 85.5 & 94.1 & 76.2 & 86.6  \\
Base + RE (double) + hard-triplet & 59.8 & 61.0 & 63.9 & 64.4 & 86.0 & 94.2 & 76.2 & 86.6  \\
% Base + RE (double) + soft-triplet & 62.1 & 63.5 & 64.5 & 66.6 & 86.1 & 94.0 & 76.1 & 86.5  \\
Base + RE$\odot$BcE + $L_{\textit{sht}}^{\textit{sub}}$ & 67.3 & 69.5 & 69.2 & 71.0  & 86.7 & 94.6 & 76.5 & 87.0  \\
%Base + BCA + hard-triplet(64+64)  & 58.9 & 61.1 &  &  &  &  &  &   \\
Base + RE$\odot$BcE + $L_{\textit{sht}}^{\textit{sub}}$+ $L_{\textit{sht}}^{\textit{full}}$  & 67.7 & 70.1 & 71.1 & 73.1 & 87.0 & 94.7 & 76.8 & 87.5  \\
%Base + BCA + triplet + Focal & - & - & 69.5 &71.1  &86.7 & 94.6 & 76.5 & 86.8  \\
%Base + BCA + hard-triplet(64+64) + Focal  & 59.5 & 60.7 & 67.3  & 68.9 & &  & 75.2 & 85.6  \\
Base + RE$\odot$BcE + $L_{\textit{sht}}^{\textit{sub}}$+ $L_{\textit{sht}}^{\textit{full}}$ +$L_{f}$  & \textbf{68.7} & \textbf{71.1} & \textbf{71.9} & \textbf{73.9} & \textbf{87.5} & \textbf{94.8} & \textbf{77.2} & \textbf{88.0} \\
% Base + BcE + hier-hard-triplet + Focal   & \textbf{68.6} & \textbf{70.9} & \textbf{70.6} & \textbf{73.7} & \textbf{87.5} & \textbf{94.8} & \textbf{77.2} &\textbf{87.8}  \\
\hline
\hline
\end{tabular}
%}
\label{tab:ab}
\end{table*}

\subsection{UMFL-enabled Single-branch Network }

\subsubsection{Experimental Settings}\label{subsec:singlebranch}

UMFL is proposed to learn multifaceted features in an unified framework to increase the generalization of CNN. Following the ASB method \cite{luo2019bag}, here the simple single-branch network ResNet-50 is used to evaluate the effectiveness of UMFL. For compound batch erasing, we first randomly choose 16 identities with four image samples per identity as a sub-batch and duplicate the sub-batch before applying any erasing operation. Then we apply BcE and RE respectively to the two sub-batches, $\mathbf{X}_{\textit{bce}}$ and $\mathbf{X}_{\textit{re}}$. The $s$ in the sub-batch $\mathbf{X}_{\textit{bce}}$ is set to a random integer in the range $[6,8]$ and the $s_{l}, s_{h}, r_{1}, r_{2}$ in the sub-batch $\mathbf{X}_{\textit{re}}$ are respectively set to $0.05, 0.4, 0.3, 0.33$ \cite{zhong2017re}. Overall, our person ReID model replaces the data augmentation and the simple loss of ASB with the compound batch erasing and the hierarchical structured loss respectively. So, our model is termed UMFL-enabled ASB below.

In this section our UMFL-enabled ASB is compared with state-of-the-art methods that use single-branch network ResNet-50, which includes four data augmentation based methods \cite{zhong2018camstyle,qian2018pose,kalayeh2018human,liu2019view} and four global feature based methods \cite{ristani2018features,huang2018adversarially,wang2018mancs,shen2018deep}. ASB is used as a baseline.
% in Section \ref{sec:augmentation} to obtain each batch $X_{full}$ with 128 batch size.

% we replace all the data augmentation (if any) of the three methods by our compound batch erasing, in which we randomly choose 16 identities of 4 images as a sub-batch and apply same operation described in Section \ref{sec:augmentation} to obtain each batch $X_{full}$ with 128 batch size. The $s$ in batch-constant erasing sub-batch is set to a random integer of $[6,8]$ and the $s_{l}, s_{h}, r_{1}, r_{2}$ in random erasing sub-batch are set to $0.05, 0.4, 0.3, 0.33$. With respect to the loss part, we replace their losses with our hierarchical structured loss in which the hyper-parameters $\gamma$ and $\alpha$ is set to $2$ and $0.1$ as the default.

\subsubsection{Results}

% To fairly and adequately evaluate the performance of our method, we first apply UMFL on ASB \cite{luo2019bag} which is a global feature based method using simplified network structure Resnet-50. We compare this UMFL-enabled ASB with several data augmentation based methods including GAN based augmentation \cite{zhong2018camstyle,qian2018pose}, segmentation based methods \cite{kalayeh2018human}, view confusion methods \cite{liu2019view}. We also compare UMFL with global feature based mehtods \cite{hermans2017defense,ristani2018features,huang2018adversarially,wang2018mancs,shen2018deep}. All these methods are based on single resnet-50 architecture with no striping. 

The performance of the single-branch network-based methods is shown in the top part in Table \ref{tab:MAP}. The results show that our method significantly outperforms all the competing methods in both mAP and R-1 on all datasets, especially on the CUHK03 detected and labeled datasets where UMFL-enabled ASB achieves more than 17.5\% improvement over the best competing methods. The CUHK03 datasets presents substantially more challenges than the other two datasets, because the number of identities there is more than that of Market1501 and DukeMTMC but the number of image samples per identity is only about half of them. UMFL excels at handling such challenging datasets, since its compound batch erasing effectively augments the data and its hierarchical structured loss can leverage these augmented data to learn diverse and discriminative features. Due to the collaboration of these two modules, the data augmentation in our UMFL-enabled ASB is more effective than the GAN, segmentation, or view confusion-based data augmentation methods \cite{zhong2018camstyle,qian2018pose,kalayeh2018human,liu2019view}. Despite the fact that ASB \cite{luo2019bag} is the most effective global feature-based method here, our UMFL-enabled ASB can still achieve substantial improvement over ASB. This is mainly because our unified multifaceted feature learning approach drives the model to attend to diverse discriminative body parts, whereas ASB mainly pays attention to only a single highly discriminative part (see Figure \ref{fig:vis} for detail).

\subsection{UMFL-enabled Striping Methods}
\subsubsection{Experimental Settings}
%Although UMFL is devised to learn multifaceted features with simple network architecture, it is applicable to other network architectures since UMFL focuses on the data augmentation and loss function modules. 
Our UMFL also can be used in the striping methods to improve the CNN's generalization. We show in this section that UMFL can be applied to substantially improve two most recent state-of-the-art striping methods, BDB\cite{dai2019batch} and ABD\cite{chen2019abd}. We only replace the data augmentation and the loss of BDB and ABD with the respective compound batch erasing and hierarchical structured loss of UMFL with all other parts fixed, which are termed UMFL-enabled BDB and ABD. Eight other state-of-the-art striping ReID methods \cite{chang2018multi,sun2018pcb,yao2019deep,hou2019interaction,zhai2019defense,li2018harmonious,li2018harmonious,xu2018attention,song2018mask} are also used as competing methods, in which \cite{li2018harmonious,xu2018attention,song2018mask} are attention based methods.

\subsubsection{Results}

The performance of the striping-based methods is shown on the lower part in Table \ref{tab:MAP}. It is clear that ABD tends to obtain better performance on Market1501 and DukeMTMC datasets but is less effective on the two challenging CUHK03 datasets. Impressively, UMFL-enabled ABD achieves 4.9\% - 5.3\% and 4.0\% - 6.0\% improvement over the original ABD on these two challenging datasets, CUHK03 detected and labeled, and obtains the best performance on all four datasets. Similarly, UMFL-enabled BDB can also substantially improve the original BDB on the Market1501 and DukeMTMC datasets that BDB performs less effectively, achieving 2.1\% - 2.7\% and 2.1\% - 2.7\% on mAP and R-1, respectively; it performs comparably to the original BDB on CUHK03 datasets.

% BDB has better performance than UMFL-enabled ASB just on CUHK03 Detected dataset. For other datasets, UMFL-enabled ASB outperforms BDB by a large margin on both mAP and R-1. Same with ABD, BDB has more parameters to learn (33M vs 25M). With our UMFL frame, though there is a slight decrease on CUHK03 detected dataset, UMFL-enabled BDB has the increase of $2.1\% - 2.7\%$ and $2.1\% - 2.7\%$ on mAP and R-1 on Market1501 and DukeMTMC datasets.

Comparing across the full table, it is remarkable that our UMFL-enabled ASB that uses a single-branch ResNet-50 backbone can outperform most striping based methods; it even performs better than BDB in 3 out of 4 datasets in both mAP and R-1. ABD generally performs better than the UMFL-enabled ASB, but it is significantly more complex method, involving nearly triple parameters than the UMFL-enabled ASB (i.e., 69M vs 25M).

%Our UMFL frame can be easy applied on different architecture. Here we use our UMFL to replace the data augmentation and loss part of striping method ABD and BDB, and a global feature based method ASB, to verify the effectiveness and universality of our UMFL frame. The results are shown on the bottom of Table. \ref{tab:MAP}. From the last three lines of the table, we can see that, the performance of the three baselines continue to grow by replacing the data augmentation and loss part with our UMFL frame. Though there is a slight decrease of Ours + BDB on CUHK03 detected dataset, our UMFL brings BDB the increase of $2.1\% - 2.7\%$ and $2.1\% - 2.7\%$ on mAP and R-1 on Market1501 and DukeMTMC datasets. With our UMFL frame ABD has $4.9\% - 5.3\%$ and $4.0\% - 6.0\%$ increase on challenge datasets, CUHK03 detected and labeled, and achieves the best performance on all datasets.

\subsection{Understanding the Effectiveness of UMFL}
\subsubsection{Ablation Study}
We evaluate the importance of three key components to UMFL, including compound batch erasing, hierarchical structured hard-triplet loss and the adapted focal loss. The ablation evaluation is performed via the UMFL-enabled ASB. The results are provided in Table \ref{tab:ab}.

\textbf{Compound Batch Erasing:RE$\odot$BcE}. The large margin of the performance between `Base + RE + hard-triplet' (the full ASB) and `Base + hard-triplet' (ASB without RE) justifies the important contribution of random erasing to the ReID performance. When we include both RE and BcE by using separate hard-triplet losses on both sub-batches, i.e., `Base + RE$\odot$BcE + $L_{\textit{sht}}^{\textit{sub}}$', we achieve significant improvement over `Base + RE + hard-triplet' across all four datasets. This is mainly due to the collaborative effect of the compound batch erasing and the separate losses on the sub-batches. As we discuss in Section \ref{subsec:sub-batch}, this collaboration enables our model to learn multifaceted discriminative features from different parts, resulting in the significantly better ReID performance.

\textbf{Hierarchical Structured Hard-triplet Loss}. We further apply the hard-triplet loss to the full batch. Combining with the hard-triplet losses on the two sub-batches, we have a hierarchical hard-triplet loss, i.e., `Base + RE$\odot$BcE + $L_{\textit{sht}}^{\textit{sub}}$+ $L_{\textit{sht}}^{\textit{full}}$'. Enforcing the same hard-triplet loss to the full batch as that in sub-batches enables the learning of finer-grained features. Thus, it can achieve consistently additional improvement over `Base + RE$\odot$BcE + $L_{\textit{sht}}^{\textit{sub}}$'.

\textbf{Adapted Focal Loss}. Lastly the adapted focal loss is added, i.e., `Base + RE$\odot$BcE + $L_{\textit{sht}}^{\textit{sub}}$+ $L_{\textit{sht}}^{\textit{full}}$ +$L_{f}$', which helps obtain further consistent improvement over `Base + RE$\odot$BcE + $L_{\textit{sht}}^{\textit{sub}}$+ $L_{\textit{sht}}^{\textit{full}}$'. This demonstrates that the adapted focal loss indeed learns additional important features from hard negative samples, for which the previous three hard triplet losses fail to do so.

\subsubsection{Visualization of Attention Maps}

\begin{figure}[t]
\begin{minipage}[b]{1\linewidth}
  \centering
  \centerline{\includegraphics[width=0.95\linewidth]{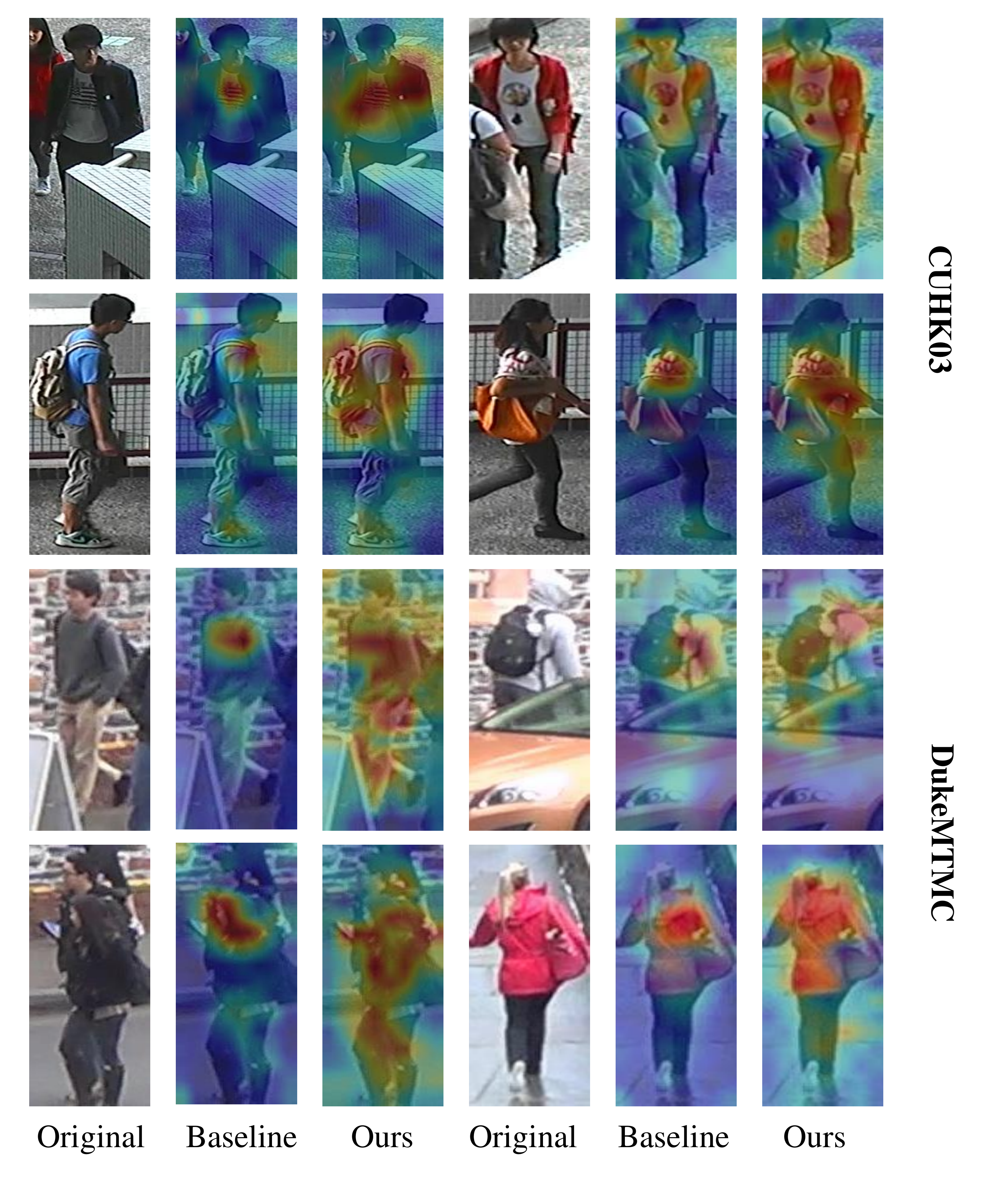}}
%  \vspace{1.5cm}
\end{minipage}
\caption{Visualization of attention maps of ASB (i.e., Baseline) and UMFL-enabled ASB (Ours).}
% It is obvious that our method learns diverse important attention while ASB focuses on small discriminative parts. }
%  As shown in column three and six, the diverse attention map from ours almost span over the whole person rather than some local areas of the baseline.
\label{fig:vis}
\end{figure}

We conduct a set of attention visualizations by using Grad-CAM visualization method \cite{selvaraju2017grad}, which output the attention map on the last output feature maps. The results of ASB \cite{luo2019bag} (Baseline) and our UMFL-enabled ASB (ours) are shown on Figure \ref{fig:vis}, from which we can see that the feature maps of ASB highlight single discriminative parts only. By contrast, the UMFL-enabled ASB can effectively ignore background or other noise information and attend to diverse discriminative parts in different cases, e.g., identity images taken different angles, identities with different accessories, and occluded identities. For example, in the 1st and 3rd rows in Figure \ref{fig:vis}, although the persons are occluded by obstacles of different size, our method can still focus on the identities and also pay attention on different parts, while ASB focuses on small highly discriminative areas only. 
% This experiment demonstrates that the proposed UMFL approach can improve attentiveness as well as the generalization ability of CNN.

% It is obvious that, the more attention on different parts of person's body, the more diverse and decorrelated of the presentation of each image on feature space. 

\subsection{Beyond Person ReID: Enabling Vehicle ReID}
To further evaluate the capability of our method, we evaluate the performance of the UMFL-enabled ASB on two vehicle ReID datasets, VeRi-776 \cite{liu2016eccv} and VehicleID \cite{liu2016deep}. VeRi-776 contains about 50,000 images of 776 vehicles across 20 cameras. VehicleID dataset contains 221,763 images with 26,267 vehicles. There are three test subsets with different sizes and we use the large test set which contains 20,038 images of 2,400 vehicles. We compare our method with six state-of-the-art vehicle ReID methods \cite{lou2019veri,khorramshahi2019dual,he2019part,luo2019bag,wang2017orientation,kanaci2018vehicle}, with ASB as the baseline. The results are shown on Table \ref{tab:MAP_Ve}. The UMFL-enabled ASB outperforms most vehicle ReID methods by a large margin. Compared to ASB, our method achieves 2.7\% improvement on mAP and and 0.7\% - 0.8\% improvement on R-1. This demonstrates that the proposed UMFL approach can effectively generalize to the vehicle ReID task.

\renewcommand{\arraystretch}{0.9}
\begin{table}[htbp]
\centering
\caption{MAP and R-1 performance on vehicle ReID datasets.}

\begin{tabular}{c|c c|c c}
\hline
\multirow{2}{*}{\textbf{Methods}}& \multicolumn{2}{ c| }{\textbf{VeRi-776}} & \multicolumn{2}{ c }{\textbf{VehicleID}}\\
& mAP & R-1 & R-1 & R-5 \\
\hline
% S-CNN \cite{shen2017learning}  & 58.3  & 83.5 & - & - \\
% VAMI  \cite{zhou2018aware}  & 50.1  & 77.0 & - & - \\
% PROVID \cite{liu2017provid}  & 53.4  & 81.6 & - & - \\
% FACT \cite{liu2016deepreid}   & 27.8  & 61.4 & - & - \\
OIFE \cite{wang2017orientation}  & 51.4  & 92.4 & 67.0 & 82.9   \\
MSVR  \cite{kanaci2018vehicle}  & 49.3  & 88.6 & 63.0 & 73.1   \\
FDA-NET \cite{lou2019veri} & 55.5& 84.3& 55.5& 74.7   \\
AAVER \cite{khorramshahi2019dual} & 66.4& 90.2& 63.5& 85.6   \\
RAM \cite{liu2018ram} & 61.5 & 88.6 & 67.7 & 84.5   \\
P-R \cite{he2019part}  & 74.3  & 94.3 & 74.2  & 86.4   \\
\texttt{ASB} \cite{luo2019bag}  & 75.7  & 95.2 & 77.5  & 91.0   \\
\texttt{UMFL-enabled ASB}   & \textbf{77.8}  & \textbf{95.9} & \textbf{78.1}  & \textbf{92.0}   \\

\hline
\hline
\end{tabular}

\label{tab:MAP_Ve}
\end{table}

\section{Conclusion}

This paper introduces a simple and effective Unified Multifaceted Feature Learning (UMFL) approach to learn diverse discriminative features expressed in different parts of identities. Learning such features often can only be achieved by using multi-branch complex networks. We show that the two key collaborative modules, batch compound erasing and hierarchical structured loss, of UMFL can effectively work together to achieve this goal using simple single-branch network only. Also, the two key modules of UMFL are generic in that (i) they can also be plugged into the complex networks to further enhance their performance in the person ReID task and (ii) they can effectively generalize to the vehicle ReID task.   

\ifCLASSOPTIONcaptionsoff
  \newpage
\fi

\bibliographystyle{IEEEtran}
\bibliography{refs}

\end{document}